\documentclass[10pt,twocolumn,letterpaper]{article}

\usepackage{times}
\usepackage{epsfig}
\usepackage{graphicx}
\usepackage{amsmath}
\usepackage{amssymb}
\usepackage[utf8]{inputenc} 
\usepackage[T1]{fontenc}    
\usepackage{url}            
\usepackage{booktabs}       
\usepackage{amsfonts}       
\usepackage{nicefrac}  
\usepackage{microtype} 
\usepackage{bm}
\usepackage{bbm}
\usepackage{times}
\usepackage{authblk}
\usepackage{epsfig}
\usepackage{algorithm}
\usepackage{algpseudocode}
\usepackage{xcolor,colortbl}
\usepackage{graphicx}
\usepackage{color}
\usepackage{stmaryrd}
\usepackage{graphics}
\usepackage{mathptmx}
\usepackage{xspace}
\usepackage{dsfont}
\usepackage{mathtools}
\usepackage{tabularx, multirow}
\usepackage{gensymb}
\usepackage{indentfirst}

\usepackage[pagebackref=true,breaklinks=true,letterpaper=true,colorlinks,bookmarks=false]{hyperref}

\begin{document}

\title{Tree-gated Deep Mixture-of-Experts For Pose-robust Face Alignment}

\author[1]{Est\`ephe Arnaud}
\author[2]{Arnaud Dapogny}
\author[1, 2]{K\'evin Bailly}

\affil[1]{\small Sorbonne Universit\'e, CNRS, Institut des Systèmes Intelligents et de Robotique, ISIR, F-75005 Paris, France}
\affil[2]{\small Datakalab, Paris, France}

\date{}
\maketitle

\begin{abstract}
	Face alignment consists of aligning a shape model on a face image. It is an active domain in computer vision as it is a preprocessing for a number of face analysis and synthesis applications. Current state-of-the-art methods already perform well on "easy" datasets, with moderate head pose variations, but may not be robust for "in-the-wild" data with poses up to $90^{\circ}$. In order to increase robustness to an ensemble of factors of variations (e.g. head pose or occlusions), a given layer (e.g. a regressor or an upstream CNN layer) can be replaced by a Mixture of Experts (MoE) layer that uses an ensemble of experts instead of a single one. The weights of this mixture can be learned as gating functions to jointly learn the experts and the corresponding weights. In this paper, we propose to use tree-structured gates which allows a hierarchical weighting of the experts (Tree-MoE). We investigate the use of Tree-MoE layers in different contexts in the frame of face alignment with cascaded regression, firstly for emphasizing relevant, more specialized feature extractors depending of a high-level semantic information such as head pose (Pose-Tree-MoE), and secondly as an overall more robust regression layer. We perform extensive experiments on several challenging face alignment datasets, demonstrating that our approach outperforms the state-of-the-art methods.
\end{abstract}

\section{Introduction}

\begin{figure*}[!t]
	\centering
	\includegraphics[width=\linewidth]{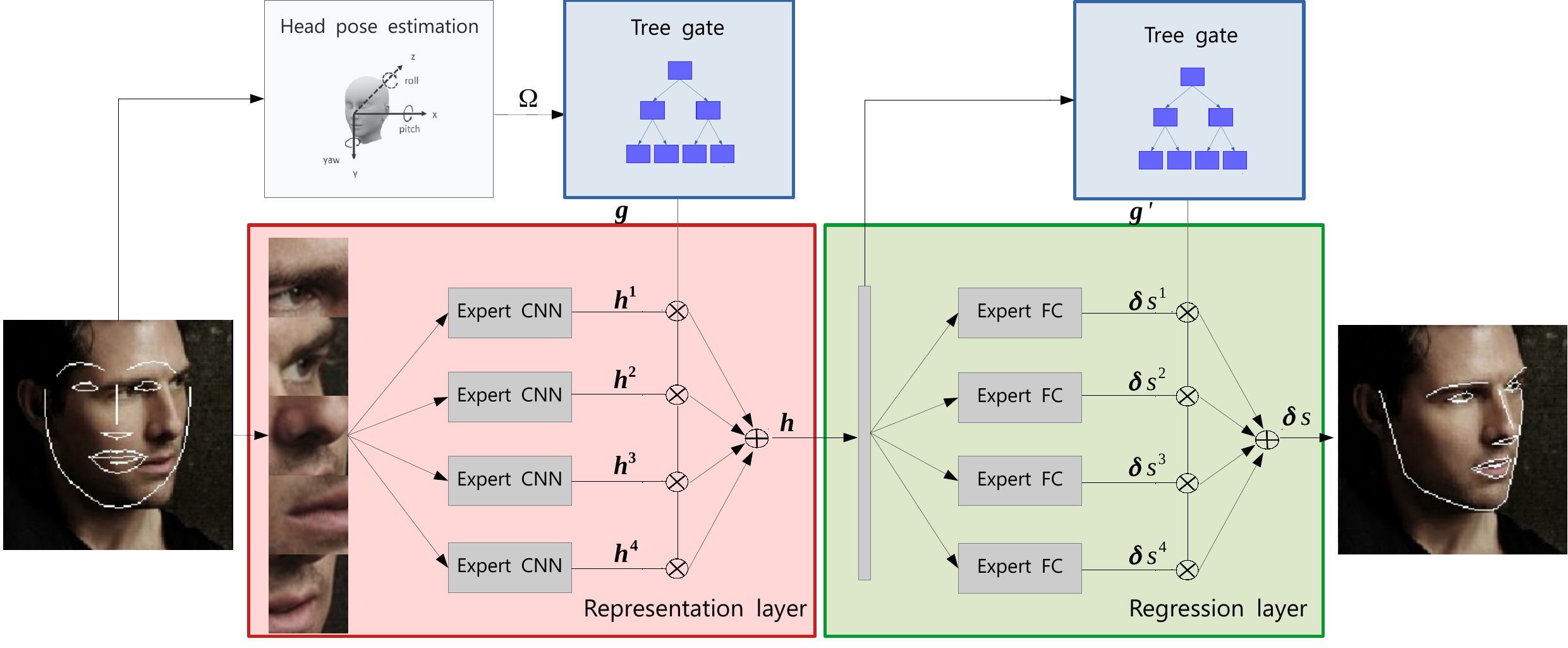}
	\caption{Architecture of the proposed Pose-Tree-MoE model for face alignment. Only one cascade stage is represented for clarity reasons. Patches are extracted around each landmark and fed to an ensemble of expert CNNs. Head pose is estimated and used to weight each expert CNN through a tree-gate. The extracted representations are then used as an input for an ensemble of expert FC layers. Contribution of each expert is, once again, weighted using a tree-gate to predict landmark displacement. These stages are then repeated through all cascade stages, allowing a pose-adaptive, robust face landmark alignment.}
	\label{overview}
	\end{figure*}

Face alignment refers to the process of localizing a number of landmarks on a face image (lips and eyes corners, pupils, nose tip). This is an important research field of computer vision, as it is an essential preprocess for applications such as face recognition, tracking, expression analysis as well as face synthesis.

Recently, regression-based methods appeared among the most successful ones, achieving high accuracies on images with reasonable variations in head pose, illumination as well as occasional facial expressions or partial occlusions. These methods address the face alignment problem by directly learning a mapping between shape-indexed face texture and the landmark positions. This regression is often performed in a cascaded manner: starting from an initial guess, a first stage predicts a displacement for every landmark coordinate. This prediction then gets progressively refined through successive regressors that are trained to improve the predictions of the previous stages. In the frame of such a coarse-to-fine strategy, the first cascade stages usually capture large deformations, while the last stages focus on more subtle variations. The work of Xiong \textit{et al.} \cite{Xiong2013} proposes to use successive linear regressions based on SIFT descriptors extracted around each landmark at the current position of the model. Similarly, Ren \textit{et al.} \cite{Ren2014} propose to learn shape-indexed pixel intensity difference features with random forests in order to speed up the feature extraction step.

Deep learning techniques have also been investigated to tackle the face alignment problem. For example, in \cite{Sun2013} each cascade stage is modeled using deep convolutional networks (CNNs) in order to jointly learn the representation and regression steps in an end-to-end fashion instead of training the regressor upon handcrafted features. Mnemonic Descent Method \cite{Trigeorgis2016} improves the feature extraction process by sharing the CNN layers among all cascade stages, and the landmark trajectories through the successive cascade stages are modeled using recurrent neural networks. This results in memory footprint reduction as well as more efficient representation learning and a more optimized cascaded alignment process.

Using deep architectures within the cascaded regression framework allows to achieve high-end alignment precision. However, despite the success of these methods, these are very sensitive to extreme conditions such as large head pose, facial expression, illumination changes, or occlusions induced by objects in front of the face (e.g. glasses, hands, hairs). The appearance of the face can then drastically change and corrupt the input features fed to the displacement regressor in the first stages of the cascades, causing errors that will be hard to overcome later on.
In order to address these limitations, ensemble methods can be combined with the use of deep learning techniques: using a committee of expert layers instead of just a single, strong layer improves the diversity of possible responses, which leads to an increased overall robustness. Furthermore, an adaptive combination of the outputs of these expert layers can be learned jointly by the use of gates. 
Using a tree gate structure allows to learn a hierarchical clustering of the expert layers, which leads to a more efficient selection and therefore specialization of the experts. These gates can be based on either an extracted representation (e.g. when applied for regression), or a high-level semantic information (such as head pose when applied for learning representation layers).

The representation layer, regression layer and the corresponding tree gates can be trained jointly in an end-to-end manner using neural trees. The proposed architecture is summarized on Figure \ref{overview}. The contributions of this paper are thus three-folds:
\begin{itemize}
\item We show that integrating ensemble methods within a deep architecture is beneficial to the overall robustness of face alignment methods. In particular, we use a committee of expert layers instead of a single, strong layer for representation and regression layer respectively. This allows each expert network to be geared towards a specific alignment case.
\item We propose an adaptive weighting strategy based on gates that learn to combine the contribution of each expert network. Weights can be estimated from a high level semantic information such as head pose, or directly from an embedding extracted by the representation layer. In addition, tree-structured gates allow to learn a set of hierarchically clustered expert layers. It can be learned as neural trees \cite{Kontschieder2016} to allow joint end-to-end training of expert networks and gates.
\item We propose a real-time face alignment system that outperforms state-of-the-art approaches on several databases and is particularly robust to large poses and occlusions.
\end{itemize}

\section{Related work}
The main current challenge of face alignment problem is the robustness to strong variations, such as large pose or occlusion. A first approach may consist in better conditioning the training in order to increase generalization capability. 
Landmark localization can be improved by simultaneously learning other related tasks such as attributes detection \cite{zhang2016learning}\cite{honari2018improving}. Indeed, learning to detect attributes such as the presence of glasses on the face can improve the model robustness to occlusions. %Moreover, jointly learning pose and landmarks localization improves the performance of both tasks. 
However, these techniques require additional data, either unlabelled or annotated with auxiliary attributes.

Other models aim to address robustness to only one source of variation. The architecture and the training procedure are then designed to specifically address this variation. For example, some models are specialized in occlusion handling and explicitly predict the occluded part of the face \cite{Burgos-Artizzu2013}\cite{Ghiasi2014}\cite{Yu2014}\cite{zhang2016occlusion}. However, learning such models generally requires data with occlusion labels, which is a major requirement.
Head pose variations can also drastically change the appearance of the face, and some landmarks can be self-occluded. Taking this information into account can improve the face alignement process. 
%Other models use head pose estimation to guide landmarks localization. Indeed, it's a highly semantic information and is particularly well adapted for face alignment: the appearance can be drastically modified for a certain variation in pose, even making some landmarks invisible in the case of large poses (more than 60\degree).
Zhu \textit{et al.} \cite{Zhu2019} proposed to integrate head pose information to condition and adapt Convolutional Neural Networks. %, and have introduced reference datasets to train and test approaches on large pose context in abundant quantities. 
Kumar \textit{et al.} \cite{Kumar2018} proposed a Bayesian formulation in which head pose estimation allows to condition heatmaps extracted to estimate landmarks localization, with the constraint that the estimated face shape must follow a dendritic structure for effective information sharing. These approaches have demonstrated that the head pose information significantly improves the localization performance. However, in this case, head pose is treated as a \textit{post-hoc} multiplicative variable. Conversely, we argue that taking this information into account upstream in the network leads to better representation learning, bringing more robustness, as head pose can dramatically affect face appearance.

Other approaches explicitly seek to build models that are specific to each variation type, and combine them together. Wu \textit{et al.} \cite{Wu2017} proposes a global framework, trained in a cascaded manner, which simultaneously performs facial landmarks localization, occlusion detection and head pose estimation with separate modules. Relationships between these allows the modules to benefit from each other. However, each module requires additional annotations in the trainset. By contrast, the proposed method only relies on facial landmarks and head pose information that can be inferred from the landmark positions. 

In addition, the architecture of our model combines advantages of ensemble methods with those of deep learning techniques. 
Such models have already been explored in the deep learning literature: a first approach for combining deep learning and ensemble methods is to craft a differentiable ensemble architecture in order to allow end-to-end parameter learning. Kontshieder et al. \cite{Kontschieder2016} designed a differentiable deep neural forest, by unifying the divide-and-conquer principle of decision trees (allowing to cluster data hierarchically) with the representation learning from deep convolution networks. Each predictor is a binary tree, whose split nodes contain routing functions, defining the probability of reaching one of the sub-trees. The probabilistic routing functions are differentiable, allowing these neural trees to be integrated into a fully-differentiable system. The forest corresponds to a set of trees, who's final output is the simple average of the output of each tree.
Since then, other models have sought to generalize neural forests \cite{Tanno2018}\cite{Ioannou2016}, by integrating upstream convolutionnal or multi-layer perceptron layers within routing functions to learn more complex input partitionings, leading to higher performance.
Dapogny et al. \cite{Dapogny2018a} used neural forests for face alignment, with promising results. However, their model uses handcrafted SIFT descriptors. In addition, to adapt neural forest for regression purposes, the authors use neural trees whose leaves are fixed and correspond to a sampling of the remaining displacements from the training data (with small variations). Their model then seeks to optimally combine fixed leaves. This training procedure can theoretically lead to rigid responses, reducing the expressiveness of the model.

A second approach may be to parallelize a set of small networks instead of a single large network. The idea of using a set of regressors within an end-to-end system was firstly introduced by Jacobs et al. \cite{Jacobs1991} and more recently taken up by Eigen \textit{et al.} \cite{eigen2013}. It shows promising results and is well adapted to our problem. Eigen \textit{et al.} design a Mixture-of-Experts (MoE) layer, consisting in jointly learning a set of expert subnetworks with gates, allowing to learn to combine a number of experts depending on the input. In the same vein, Shazeer \textit{et al.} \cite{shazeer2017outrageously} introduce sparsity in MoE in order to save computation and to increase representation capacity.

\section{Framework overview}
In this section, we introduce our Pose-Tree-MoE model for face alignment: first, in Section \ref{hpe} we describe the head pose estimation module. We then detail in Section \ref{feat} the representation layer, which select relevant experts based on this head pose estimate, and extract features from patches extracted around a current feature point localization. Then, Section \ref{pred} shows how we predict landmark coordinate displacements from this pose specific representation. In Section \ref{gating} we detail the architecture of the gates used to weight the contribution of each expert network for both the representation and regression layer respectively. The whole architecture is integrated into a cascaded regression framework \ref{models}. This section also provides implementation and architectural details to ensure reproducibility of the results.

\subsection{Head pose estimation}\label{hpe}
\begin{figure}[!ht]
	\centering
	\includegraphics[width=\linewidth]{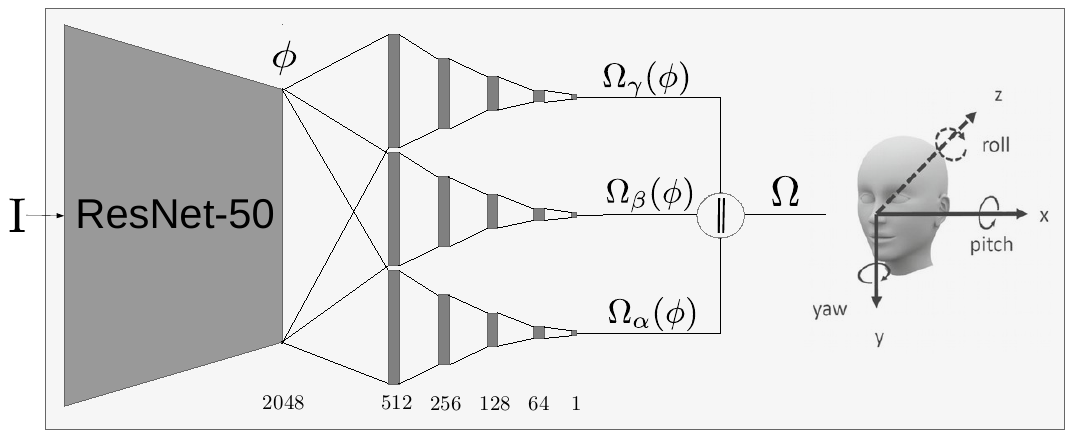}
	\caption{Overview of head pose estimator. Head pose is estimated by three distinct regressors stacked on a common representation extracted by a ResNet-50.}
	\label{pose}
\end{figure}
Following \cite{Ruiz2018} we use a truncated pre-trained ResNet-50 network to extract head pose from the raw face image $I$. We note $\phi$ the embeddings ($2048$ units) of the last fully-connected layer. A naive approach would consist in using a single deep network $\Omega(\phi(I))$ that directly predicts the 3 head pose angles, as it was done in \cite{Ruiz2018}. However, as pointed out in \cite{misra2016cross}, sharing all the representations layers may or may not be optimal, depending on the tasks at stake. In our case, we obtained better performance by regressing yaw $\gamma$, pitch $\beta$ and roll $\alpha$  values with separate networks:
\begin{equation}
\Omega = \Omega_{\gamma}(\phi) || \Omega_{\beta}(\phi) || \Omega_{\alpha}(\phi)
\end{equation}
with $\Omega_{\gamma}(\phi)$, $\Omega_{\beta}(\phi)$, $\Omega_{\alpha}(\phi) \in [-\pi, \pi]^3$ and $||$ the concatenation operator. Lastly, we also fine-tune the ResNet-50 backbone to improve the head pose regression accuracy.

\begin{figure*}[!ht]
	\centering
	\includegraphics[width=\linewidth]{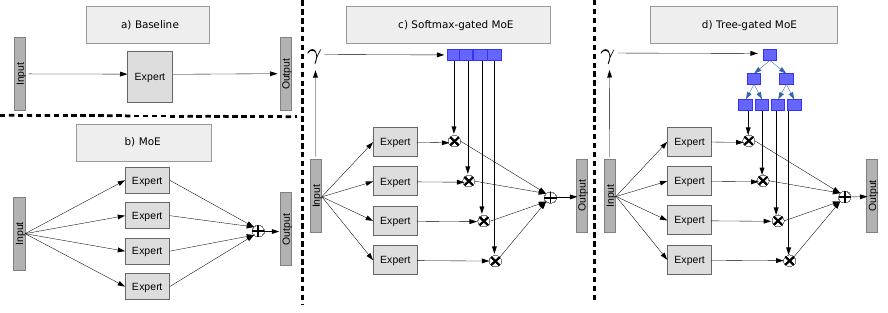}
	\caption{Architecture overview. \textbf{a) Baseline}: a baseline, single-expert network. \textbf{b) MoE}: the output is the average of $L$ independent experts. \textbf{c) Softmax-gated MoE}: The output of $L$ expert networks are weighted by the output of a $L$-dimensional softmax layer. \textbf{d) Tree-gated MoE}: The output of $L$ expert networks are weighted by the terminal (leaf) probabilities of a $\log_2(L)$-deep neural tree. The function $\gamma$ allows to gate expert networks with a transformation of the input. In our case, for the representation layer, $\gamma$ is an head pose estimate computed on the whole image. For the regression layer, $\gamma$ is the identity mapping.}
	\label{architectures}
\end{figure*}

\subsection{Representation layer}\label{feat}
\begin{figure}[!ht]
	\centering
	\includegraphics[width=\linewidth]{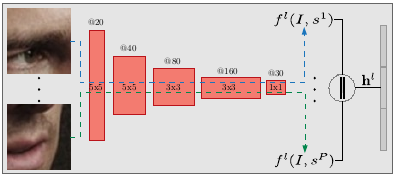}
	\caption{Overview of an expert CNN in representation layer. For each cascade stage, patches are extracted around feature points and a number of shared convolution kernels are applied to each patch. The output features are then concatenated.}
	\label{kernels_sharing}
\end{figure}

Let's note $\textbf{s} \in \mathbb{R}^{2P}$ an initial shape guess (with $P$ landmarks). As it is classical in the frame of cascaded regression for face alignment, we extract shape-indexed patches, \textit{i.e.} patches centered at each landmark current localization. If we denote $\textbf{h}$ the output of the feature extraction layer, we thus have:

\begin{equation}
    \textbf{h}=f(I,s^1)||\dots||f(I,s^{P})
\end{equation}

As it is ubiquitous among recent face alignment methods, we use CNNs to model function $f$. Furthermore, in order to limit the number of parameters, we share the convolution kernels between the different patches, as it was done in \cite{Trigeorgis2016}. Instead of using one large CNN to extract features around each landmark current location estimation \cite{Arnaud2019}, we use a committee of smaller expert CNNs. The idea is that head poses or partial occlusions can dramatically alter the appearance around specific landmarks (e.g. cheeks landmarks in case of large poses) and that we can learn expert CNNs to extract more relevant shape-indexed features in such cases. Specifically, we define $L$ expert CNN $\{f^l(I,s^1)\}_{l=1 \dots L}$ and the output of each expert CNN $\textbf{h}^l$ as follows:

\begin{equation}
    \textbf{h}^l=f^l(I,s^1)||\dots||f^l(I,s^{P})
\end{equation}

with $\textbf{h}^l \in \mathbb{R}^{Pn}$ and $n$ the number of features per landmark at the output of the last CNN layer. We denote $\textbf{H}= [\textbf{h}^1, \dots ,\textbf{h}^{L}]$ the responses of all the expert CNNs. More precisely, $\textbf{H}$ is a $Pn \times L$ matrix whose columns are the responses of the $L$ expert CNNs. Now that we have extracted expert features relatively to the neighbouring of each landmark, we want to aggregate these features: to do so a naive approach would be to simply sum these features, i.e. sum over all the columns of $\textbf{H}$. However, a better solution would be to use a high level semantic variable, such as head pose, to select the most relevant experts based on the output of a gate function $g : \Omega \in [-\pi, \pi]^3 \mapsto \textbf{g}(\Omega) \in [0, 1]^L$ such that $\sum_{l=1}^L g_l(\Omega)=1$. In such a case, the output \textbf{h} of the representation layer can be written as the sum of the contributions of the $L$ experts, weighted by the gate value relatively to that expert:

\begin{equation}
    \textbf{h} = \textbf{H}.\textbf{g}(\Omega)
\end{equation}

\subsection{Regression layer}\label{pred}
Given the extracted feature vector $\textbf{h} \in \mathbb{R}^{Pn}$, we now aim at regressing a displacement $\delta \textbf{s} \in \mathbb{R}^{P}$ between $\textbf{s}$ and the ground truth landmark localization $\textbf{s}^*$. Such displacement is usually estimated using a single, large deep fully-connected network, \textit{i.e.} $\delta \textbf{s} = r(\textbf{h})$.

Once again, instead of designing one such large network, we can use a committee of $L'$ several smaller expert networks %$\delta \textbf{s}^1, \dots, \delta \textbf{s}^{L'}$.
$\left\{ \delta \textbf{s}^l = r^l(\textbf{h}) \right\}_{l=1\ldots L'}$. Let's note $ \delta \textbf{S} $ a $\mathbb{R}^{2P \times L'}$ matrix containing all the predictions of these expert regressors:

\begin{equation}
    \delta \textbf{S} = [\delta \textbf{s}^1, \dots, \delta \textbf{s}^{L'}]
\end{equation}

The columns of $\delta \textbf{S}$ contain displacements predicted by each expert regressor. More specifically, each displacements predicted by each expert indexed by $l'$ corresponds to the output of a fully connected layer with ReLU activation:

\begin{equation}
     \delta \textbf{s}^l = w_1^l . \max(0, \textbf{w}_0^l . \textbf{h}+b_0^l) + b_1^l
\end{equation}

with $ \Theta_{r}^l = \{\textbf{w}_0^l, b_0^l, w_1^l, b_1^l\} $ the set of parameters of the $l'$-th expert. Let's suppose we now have access to the output of a gating network $g' : \textbf{h} \in \mathbb{R}^{L} \mapsto \textbf{g}' \in [0, 1]^{L'}$ based on the extracted features $\textbf{h}$. In such a case, The output of the regression layer can be written as:

\begin{equation}
     \delta \textbf{s} = \delta \textbf{S} .\textbf{g'}(\textbf{h})
\end{equation}

\subsection{Gating network}\label{gating}
In what follows, we note $\gamma : \textbf{z} \in \mathbb{Z} \mapsto \textbf{x} \in \mathbb{X}$ any mapping function and $g : \textbf{x} \in \mathbb{X} \mapsto \textbf{g} \in [0, 1]^L$ a gate function with $L$ the number of expert networks associated to this gate, such that $\sum_{l=1}^L g_l(x)=1$. In order to learn an adaptive combination of expert networks, two types of gates can be designed.

\subsubsection{Softmax gate}

The most straightforward way to design a gating function is to use a softmax activation function:

\begin{equation}
     \textbf{g}(\textbf{x}) = softmax(\textbf{x}) = (\frac{e^{\textbf{w}_1.\textbf{x} + b_1}}{\Sigma_{l = 1}^L e^{\textbf{w}_l.\textbf{x} + b_l}}, \dots, \frac{e^{\textbf{w}_L.\textbf{x} + b_L}}{\Sigma_{l = 1}^L e^{\textbf{w}_l.\textbf{x} + b_l}})
\end{equation}
    with $\Theta_{g}= \{\textbf{w}_l, b_l\}_{l \in \{1, \dots, L\}}$ the parameters of the gate function. While this function is very simple, it doesn't allow to learn hierarchical partitions of $\mathbb{X}$.

\subsubsection{Tree gate}

    In order to learn a more potent gating network we use a single neural tree. A neural tree \cite{Kontschieder2016} is composed of subsequent soft, probabilistic routing functions $d_n$, that represents the probability to reach the left child of node $n$. Formally, $d_n$ is defined as a single neuron:
    \begin{equation}
        d_n(\textbf{x}) = \sigma(\textbf{w}_n.\textbf{x} + b_n) = \frac{e^{\textbf{w}_n.\textbf{x} + b_n}}{1 + e^{\textbf{w}_n.\textbf{x} + b_n}}
    \end{equation}

    with $\Theta_g = \{\textbf{w}_n, b_n\}_{n \in \mathcal{N}}$ parameters to learn.

    For an input $\textbf{x}$, the probability $\mu_l$ to reach a leaf $l \in \{1,\dots,L\}$ is computed as a product of the successive activations $d_n$ down the whole tree. Therefore:
    \begin{equation}
      \mu_l(\textbf{x}) = \prod_{n \in \mathcal{N}}d_n(\textbf{x})^{\mathds{1}_{l \swarrow n}}(1 - d_n(\textbf{x}))^{\mathds{1}_{l \searrow n}}  
    \end{equation}
    where $l \swarrow n$ is true if $l$ belongs to the left subtree of node $n$, and $l \searrow n$ is true if $l$ belongs to the right subtree.

    We define our tree-gate as the concatenation of the $2^\mathcal{D}$ leaves probabilities of a single neural tree of depth $\mathcal{D}$:
    \begin{equation}
        \textbf{g}(\textbf{x}) =  \mu_{1}(\textbf{x}) || \dots || \mu_{L}(\textbf{x})
    \end{equation}

    For extracting representations with a committee of experts layers, a naive solution would be to use the raw image as the gate input, \textit{i.e.} setting $\gamma=Identity$. However, in this case, the raw image is too low-level and cannot be used directly, thus it is preferable to use high-level semantic information computed from $I$, such as head pose estimation, by setting $\gamma=\Omega$ (see Section \ref{hpe}). By contrast, in the regression layer, the information extracted by the representation layer is already semantically abstract, thus it can directly be used as the gate input, by setting $\gamma = Identity$. %%For gating the expert regressors, however, we use $\gamma=Identity$. 
    Last but not least, the feature representation, regression and the gate parameters are all optimized jointly for each cascade stage.

\subsection{Architectures} \label{models}

Similarly to other cascaded regression approaches, we learn a cascade of mappings $\textbf{s}^{(0)} + \sum_k \delta \textbf{s}^{(k)}$, where $\textbf{s}^{(0)} \in \mathbb{R}^{2P}$ is an initial guess (usually an average shape computed over the whole train set) and each $\textbf{s}^{(k)}$ is a displacement estimated by first computing representations from shape-indexed patches centered at the current landmark estimate (as provided by the displacements applied so far) for each expert CNN, and using head pose estimate to weight the expert CNNs according to their relevance \textit{via} the gating function. From these representations, we compute the landmark displacement, once again by using a gated mixture of experts. Then, the landmark localization is updated and the subsequent representation/regression stages can be applied sequentially.

\vspace{0.25cm}
\noindent
\textbf{Regarding the regression layer:} we define several architectures whose differences only lie in the architecture of the regression layer. In such a case, we use a single CNN ($ L=1, g^l=1 \forall l $) for the representation layer of each cascade stage $ k $, which is composed of $ 5 $ strided convolution layers ($ 20 \verb+@+ 5 \times 5 \rightarrow 40 \verb+@+ 5 \times 5 \rightarrow 80 \verb+@+ 3 \times 3 \rightarrow 160 \verb+@+ 3 \times 3 \rightarrow 30 \verb+@+ 1 \times 1 $), which makes $ 30 $ features per landmark in the last layer. The output concatenated feature vector \textbf{h} is then of size $ 30 \times P $ (2040 with $ P = 68 $ landmarks). The different architectures proposed for modelling the regression layer are then as follows.

\begin{itemize}
    \item \textbf{Baseline}: a single large FC ($L'=1, g'^l=1$ $\forall l$) with one hidden layer $2040 \times 8192 \rightarrow 8192 \times136$

    \item \textbf{Mixture-of-Experts (MoE):} an unweighted combination of small expert FC: $ \textbf{g}' = (\frac{1}{L'}, \dots, \frac{1}{L'})$. We use $L'=64$ experts, each containing one hidden layer $2040\times 128 \rightarrow 128 \times 136$.

    \item \textbf{Softmax-gated Mixture-of-Experts (Softmax-MoE):} a committee of $L'=64$ expert FC layers, each one containing a single hidden layer $2040\times 128 \rightarrow 128\times136$ gated by a single 64-dimensional softmax layer.

    \item \textbf{Tree-gated Mixture-of-Experts (Tree-MoE):} a committee of $L'=64$ experts, each containing one hidden layer $2040\times 128 \rightarrow 128\times136$, gated by neural tree of depth 6 (with $2^6=64$ leaves).

\end{itemize}

\vspace{0.25cm}
\noindent
\textbf{Regarding the representation layer:} furthermore, instead of using a single deep CNN for learning representations, we use a committee of $ L_h = 8 $ expert CNNs, each being composed of $ 5 $ strided convolution layers with fewer feature maps but as many features per landmark in the last layer ($7 \verb+@+ 5 \times 5 \rightarrow 14 \verb+@+ 5 \times 5 \rightarrow 28 \verb+@+ 3 \times 3 \rightarrow 56 \verb+@+ 3 \times 3 \rightarrow 30 \verb+@+ 1 \times 1 $). We use a tree-gate based on the previously computed head pose estimate as the gating network for learning representations and refer to this model as \textbf{Pose-Tree-MoE}. We can also use softmax-gate and refer to this model as \textbf{Pose-Softmax-MoE}.

\vspace{0.25cm}
\noindent
\textbf{Implementation details:} with this configuration, each model has roughly the same total number of parameters (18 million parameters total for each cascade stage and with representation/regression layers), allowing a fair comparison between the models. Training is done by optimizing a $\mathcal{L}_2$-loss with $150 \times 150$ grayscale images and $32 \times 32$ patches centered around the 68 landmarks. The images are normalized so that they take values in $[-1, 1]$. We train $4$ cascade stages and apply data augmentation as it is traditionally done in the literature: for each image we augment the initial mean shape by a random translation factor $t \sim \mathcal{N}(0, 10)$ and a random scaling factor $s \sim \mathcal{N}(1, 0.1)$, and half the time a horizontal flip of the image is performed. The parameters (for the representations and regression layers, as well as the gating function) are optimized jointly in an end-to-end manner by applying ADAM optimizer \cite{Zhou2017} with a learning rate of 0.001.

\section{Experiments}
In this section, we validate our model both qualitatively and quantitatively. First, in Section \ref{datasets} we present the datasets that we use to train or test the proposed approaches.
We validate the architectural choices in Section \ref{archicomparison} on frontal head poses. We then compare our model with state-of-the-art approaches in Section \ref{compsota} for both 2D and 3D face alignment. In Section \ref{visu}, we qualitatively assess the relevance of the proposed approach both for the representation layer and in the regression layer. Finally, in Section \ref{runtime} we evaluate the runtime of our model.

\subsection{Datasets}\label{datasets}
We evaluate the effectiveness of the proposed approach both for 2D and 3D face alignment. In the first case, the ground truth landmarks correspond to projections on the visible part of the face. In the latter case, the ground truth corresponds to the real 2D coordinates (without the depth component) of the landmarks, which are often occluded due to large pose variations.

\subsubsection{2D Face alignment}
The 300W dataset was introduced by the I-BUG team \cite{Sagonas2015} and is considered the benchmark dataset for training and testing face alignment models, with moderate variations in head pose, facial expressions and illuminations. It also embraces a few occluded images. The 300W dataset consists of four datasets: LFPW (811 images for train / 224 images for test), HELEN (2000 images for train / 330 images for test), AFW (337 images for train) and IBUG (135 images for test). As it is classically done in the litterature for 2D face alignment, we train our models on a concatenation on AFW, LFPW and HELEN trainsets, which makes a total of 3148 images for train. For comparison with state-of-the art methods, we refer to LFPW and HELEN test sets as the common subset and I-BUG as the challenging subset of 300W, as it is commonly done in the literature.

The COFW dataset \cite{Burgos-Artizzu2013}, is an ”in-the-wild” dataset containing only occluded data. It is a benchmark dataset to test the robustness of models w.r.t. partial occlusions. COFW contains 500 images for train and 507 images for test. The models are trained with 68 landmarks annotated for each image of the datasets. However, COFW only contains images with 29 annotated landmarks. Thus, we use the method proposed in \cite{Ghiasi2014} to perform a linear mapping between the predictions made on the 68 landmarks to the 29 landmarks, as it is a common practice on this dataset.

For 2D face alignment, the evaluation metric used is the normalized mean error (NME), corresponding to the average point-to-point distance between the ground truth and the predicted shape, normalized by the inter-pupil distance, as it is classically done in the literature:
\begin{equation}
    NME = \frac{1}{N} \sum_{i=1}^{N} \frac{|| \hat{\textbf{s}}_i - \textbf{s}_i ||_2}{|| \textbf{g}_{i, l} - \textbf{g}_{i, r} ||_2}
\end{equation}
where $\hat{\textbf{s}}_i$ the prediction, $\textbf{s}_i$ the ground truth, and $\textbf{g}_{i, l}$, $\textbf{g}_{i, r}$ the left and right pupil centers respectively.

\subsubsection{3D Face alignment}
\label{description_aflw}
The 300W-LP database is a large-pose dataset synthethized from 300W, and contains face images with large variations in pose on the yaw axis, ranging from $-90\degree$ to $+90\degree$. The database contains a total of $61225$ images obtained by generating additional views of the images from AFW, LFPW, HELEN and I-BUG, using the algorithm from \cite{Zhu2016}. As it is done in state-of-the-art approaches \cite{Zhu2019}, we train on the augmented images corresponding to 300W trainset as well as their flipped counterparts, making a total of 101144 images for train.

The AFLW2000-3D dataset consists of fitted 3D faces and large-pose images for the first 2000 images of the AFLW database \cite{kostinger2011}. As it was done in \cite{Zhu2019}, we evaluate the capacities of our method to deal with non-frontal poses by training on 300W-LP and testing on AFLW2000-3D. This database consists of $1306$ examples in the $[0, 30]$ absolute degree yaw range, 462 examples in the $[30, 60]$ range and 232 examples in the $[60, 90]$ range. As in \cite{Zhu2019}, we report accuracy for each pose range separately, as well as the mean across those three pose ranges.

3D face alignment consists in localizing the $(x, y)$ coordinates of the ``true'' landmarks, as opposed to 2D alignment in which the landmarks are projected on the visible part of the face (e.g. cheeks in case of rotations around the yaw axis). In such a case, the evaluation metric used is also the normalized mean error, but the normalization is the size of ground truth bounding box, as it is introduced in \cite{Zhu2019}:
\begin{equation}
    NME = \frac{1}{N} \sum_{i=1}^{N} \frac{|| \hat{\textbf{s}}_i - \textbf{s}_i ||_2}{\sqrt{h_i \times w_i}}
\end{equation}
where $h_i$, $w_i$ the height and width of the face bounding box, respectively.

% \subsection{Head pose estimation}
% In order to be able to route the features committee into the face alignment problem according to head pose,  it is necessary to build a model capable of estimating its components.

% \subsubsection{Single/multi head.s methods comparison}
% In the same vein as \cite{deepheadpose}, we learn a regressor stacked on fine-tuned ResNet-50 to estimate head pose. We propose two architectures for the regressor with a constant amount of parameters to learn:
% \begin{itemize}
    %\item A single head corresponding to an FC containing a "large" hidden layer, and the output are pose components.
    %\item Three heads where each corresponds to an FC containing a "small" hidden layer, and the output is one of the pose components.
%\end{itemize}

%In the following, head pose estimator is trained from the dataset which is then used to learn alignment (\textit{e.g.} T-dMoE learning on 300W(-LP) will use head pose estimator also learned on 300W(-LP)).

% QQ-plot de YPR avec 1T, et 3T sur 300W et 300WLP (3x2x2)
% QQ-plot de YPR avec 300W, et 300WLP. (3x2)
% Images de prédiction (3x3)

% \subsubsection{Comparison with state-of-the-art approaches}

\subsection{Architectures comparison}\label{archicomparison}

First, we compare the different architectures detailed in Section \ref{models}. For a fair comparison, all models have roughly the same number of parameters. Pose-Tree-MoE and Pose-Softmax-MoE use a pretrained head pose estimation model. Once head pose is predicted, each of these models uses it to gate the representations extracted by the committee.
The other models proposed in comparison (Baseline, MoE, Softmax-MoE, Tree-MoE) do not use head pose estimation. The results are shown on Table \ref{ensemblecomp}.

\begin{table}[!h]
	\centering
	\caption{Comparison of different architectures in term of NME ($\%$).}
	\begin{tabular}{c|c|c|c|c}
		\hline
		Method&LFPW&HELEN&I-BUG&COFW\\
		\hline
		\hline
		Baseline & 3.92 & 4.42 & 8.89 & 5.9 \\
		MoE & 3.79 & 4.25 & 8.95 & 5.87 \\
		Softmax-MoE & 3.74 & 4.2 & 8.8 & 5.84 \\
		Tree-MoE & \textbf{3.74} & 4.2 & 8.38 & 5.76 \\
		Pose-Softmax-MoE & 3.84 & 4.27 & 7.99 & 6.12\\
		Pose-Tree-MoE & 3.82 & \textbf{4.17} & \textbf{7.5} & \textbf{5.58} \\
		\hline
	\end{tabular}
	\label{ensemblecomp}
\end{table}

In particular, the non gated regressor ensemble is more robust than a single regressor: performance is improved by 3.8\% on 300W-Common (LFPW + HELEN).
Moreover, adding gates improves performance, especially with tree-gates.
Robustness to pose is improved thanks to softmax-gates ($8.95 \rightarrow 8.8$ on I-BUG) and significantly improved thanks to tree-gates ($8.95 \rightarrow 8.38$ on I-BUG).
Robustness to occlusions is slightly improved thanks to softmax-gates ($5.87 \rightarrow 5.84$ on COFW) and significantly improved thanks to tree-gates ($5.87 \rightarrow 5.76$ on COFW).
This shows that using tree-gated ensembles of regressors allows to substantially increase the overall robustness of the model, particularly in the case of partial occlusions.

Furthermore, using head pose to gate MoE CNN models (Pose-Softmax-MoE and  tree-gated-MoE) allows to significantly increase the alignment accuracy on I-BUG, which contains several examples of non-frontal head poses. This is however not the case for Pose-Softmax-MoE model on COFW database, which contains occluded examples. By contrast, Pose-Tree-MoE model generalizes better on COFW ($5.76 \rightarrow 5.58$), and I-BUG ($8.38 \rightarrow 7.5$), without significantly degrading performance on frontal faces ($4.01 \rightarrow 4.03$ on average on LFPW and HELEN testsets).

These results show that using ensemble of experts allows for greater robustness, for modelling both the regression and representation layers. The use of gates also allows each expert to be more specialized for a given representation, leading to greater robustness. Last but not least, the hierarchical aspect of tree-gates further improves the use of expert regressors. Conditioning the learned representation to head pose estimation and taking advantage of using ensemble methods all the while learning the gates and expert layers jointly allows these experts to better co-adapt, leading to maximum robustness and accuracy.

\subsection{Comparison with state-of-the-art approaches}\label{compsota}
\begin{figure*}[!ht]
	\centering
	\includegraphics[width=\linewidth]{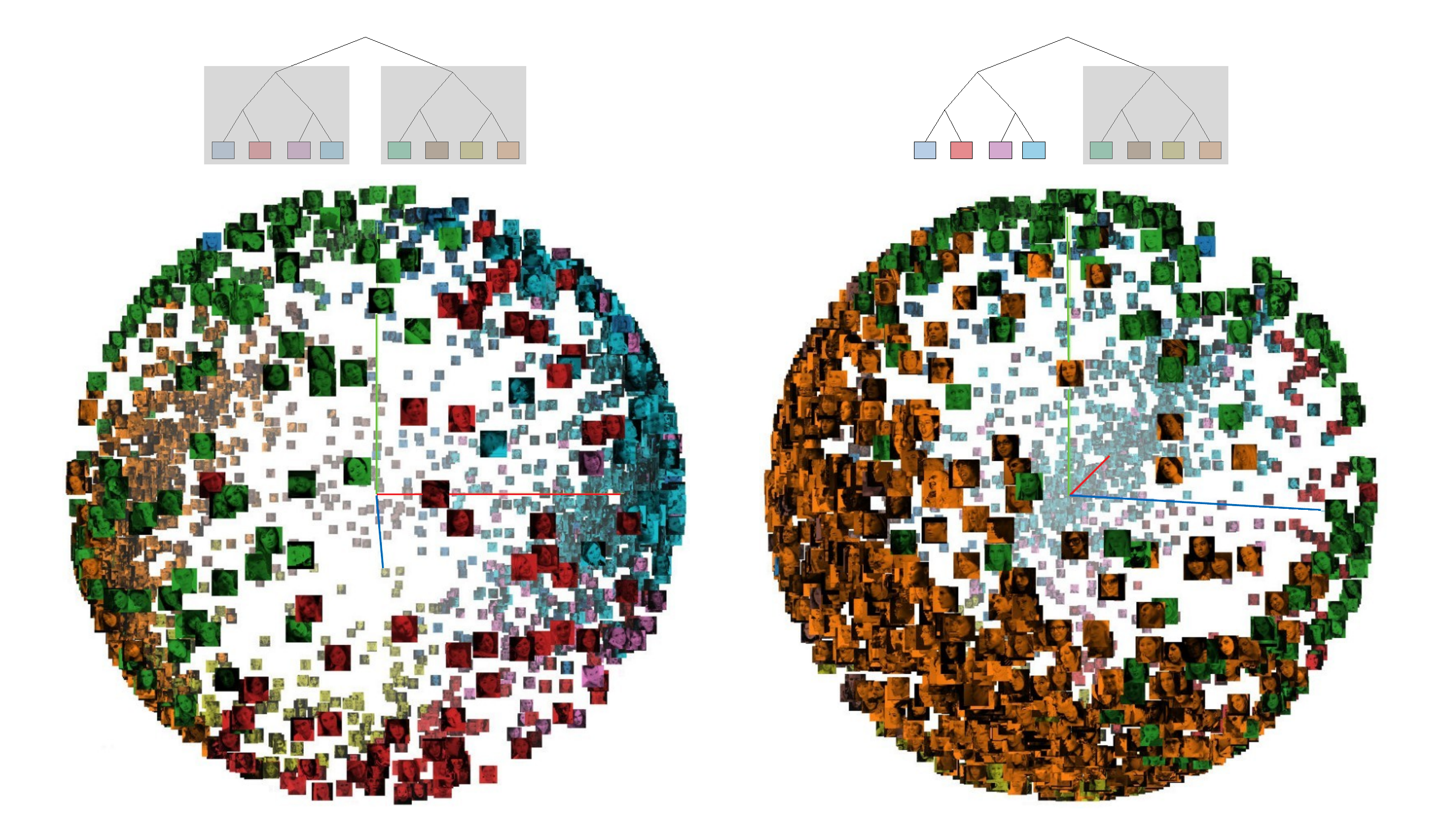}
	\caption{Visualization of representation clusters (Pose-Tree-MoE on AFLW2000-3D). \textbf{Top}: Associated color for each cluster (leaf of the 3-depth tree). Each image is then colored according to the cluster with the highest weight in the final output. \textbf{Bottom}: Dispersion in head pose space (red axis: yaw, green axis: pitch, blue axis: roll). For more visibility, the data is normalized by shifting each point by the centroid and making it unit norm. At the first level of the tree, the split is mainly due to the yaw orientation, as illustrated by the comparison of the purple/blue images in the left graph with the orange/brown images in the right graph. For the right sub-tree, the split mainly uses the yaw and roll intensities, as shown on the right graph.}
	\label{clusters}
\end{figure*}

\subsubsection{2D Face alignment}
\begin{table}[!ht]
	\centering
		\caption{Comparison with state-of-the-art approaches in term of NME (\%).}
		\begin{tabular}{c|c|c|c}
			\hline
			Method   & Common & Challenging & COFW \\
			\hline
			\hline
			RCPR \cite{Burgos-Artizzu2013} & 6.18  & 17.3 & 8.50\\
			SDM \cite{Xiong2013}           & 5.57  & 15.4 & 7.70\\
			PIFA \cite{jourabloo2017pose}  & 5.43 & 9.98 & -\\
			LBF \cite{Ren2014}             & 4.87  & 11.98 & 13.7\\
			TCDCN \cite{zhang2016learning} & 4.80 & 8.60 & -\\
			CSP-dGNF \cite{Dapogny2018a}   & 4.76  & 12.00  & - \\
			RAR \cite{xiao2016robust}      & 4.12 & 8.35 & 6.03 \\
			RCN$^+$ \cite{honari2018improving} & 4.20 & 7.78 & -\\
			DRDA \cite{zhang2016occlusion} & - & 10.79 & 6.46\\
			SFLD \cite{Wu2017} & - & - & 6.40\\
			%3DDFA (trained on 300W-LP) \cite{Zhu2019} & 5.09 & 8.07 & -\\
			PCD-CNN \cite{Kumar2018} & \textbf{3.67} & 7.62 & 5.77\\
			\hline
			\hline
			Pose-Tree-MoE & 4.03  & \textbf{7.5} & \textbf{5.58} \\
			\hline
		\end{tabular}
	\label{t_300w}
\end{table}

Table \ref{t_300w} shows a comparison between our approach and other recent state-of-the-art methods on both 300W (common and challenging subsets) and COFW databases. Our model outperforms these approaches on both 300W and COFW databases. The results on COFW show the robustness of our model to occlusions. The alignment error is similar to the human performance on this dataset ($5.60$ \cite{Burgos-Artizzu2013}).
 PCD-CNN \cite{Kumar2018} essentially uses head pose estimation as a multiplicative variable in a post-hoc processing fashion. By contrast, in Pose-Tree-MoE, head pose is used to select more relevant specialist CNNs to extract adequate features for each head pose range. As one can see, while PCD-CNN is better on 300W-Common, Pose-Tree-MoE significantly outperforms it on both 300W-Challenging and COFW. Therefore, using ensemble of tree-gated experts appears as a more robust way to adapt a face alignment network using head pose information, that leads to an overall better robustness to large variations in the data.

\subsubsection{3D Face alignment}
\begin{table}[!ht]
	\centering
		\caption{Comparison with state-of-the-art approaches on AFLW2000-3D in term of NME (\%) for several yaw ranges.}
		\begin{tabular}{c|c|c|c|c}
			\hline
			Method   & $[0, 30]$ & $[30, 60]$ & $[60, 90]$ & Mean \\
			\hline
			\hline
			LBF \cite{Ren2014}             & 8.15  & 9.49 & 12.91 & 10.19 \\
			ESR \cite{Cao2014}             & 4.60 & 6.70 & 12.67 & 7.99 \\
			CFSS \cite{zhu2015}             & 4.77 & 6.71 & 11.79 & 7.76 \\
			RCPR \cite{Burgos-Artizzu2013}             & 4.26 & 5.96 & 13.18 & 7.80 \\
			MDM \cite{Trigeorgis2016}             & 4.85 & 5.92 & 8.47 & 6.41 \\
			SDM \cite{Xiong2013}             & 3.67 & 4.94 & 9.76 & 6.12 \\
			3DDFA \cite{Zhu2019}             & 2.84 & \textbf{3.57} & 4.96 & \textbf{3.79} \\
			\hline
			\hline
			Tree-MoE & 2.84 & 4.01 & 4.93 & 3.92 \\
			Pose-Tree-MoE            & \textbf{2.78} & 3.97 & \textbf{4.76} & 3.84 \\
			\hline
		\end{tabular}
	\label{t_300wlp}
\end{table}

Table \ref{t_300wlp} shows a comparison between our approach and other recent state-of-the-art methods for 3D face alignment on AFLW2000-3D. The state-of-the-art is achieved by the extended version of 3DDFA \cite{Zhu2019}, which fits a 3D dense face model before estimating a sparse set of 68 landmarks. Our Pose-Tree-MoE achieves similar performance as compared to 3DDFA \cite{Zhu2019}, all the while substantially outperforming it on large head poses in the $[60, 90]$ range. Our model also significantly outperforms all other state-of-the-art approaches on this dataset \cite{Ren2014,Cao2014,zhu2015,Burgos-Artizzu2013,Trigeorgis2016,Xiong2013}. Furthermore, contrary to \cite{Zhu2019}, our approach only aligns a sparse set of landmarks, thus only requires ground truth landmarks for training, as opposed to the parameters of a morphable model. This shows that using a tree-gated committee of expert CNNs allows to learn relevant experts for each pose range, that produce suitable representations upon which the tree-gated MoE layer can adaptively align the facial landmarks.
Conditioning the representation using the head pose estimate significantly improves the results on large poses. This is confirmed by the comparison between Pose-Tree-Moe and Tree-MoE.

In what follows, we propose a number of qualitative experiments to assess that the head pose clustering of the expert CNNs behaves as expected.

\subsection{Qualitative evaluation}\label{visu}
In this section, we conduct some experiment to provide insight on how the gated models behave, by visualizing the contributions of tree-gates:
\begin{itemize}
    \item Interpretability through hierarchical clustering visualization in representation layer. This allows to study how the model splits the poses space in order to extract the representation. In addition, this ensures consistency between the spatial distribution of poses and the use of expert CNNs.
    \item Efficiency through the distribution and use of expert FCs in regression layer. This allows for fewer regressors to be used, whose predictions are accurate.
\end{itemize}

\subsubsection{Representation layer}
As seen in the previous section, integrating head pose information to extract representations significantly improves the robustness to strong variations in pose, and results in a model that exceeds the state-of-the-art.
It might be interesting to introspect the model in order to study how it behaves. To do this, we propose to visualize hierarchical clustering performed by our model on a dataset with maximum pose variability, such as AFLW2000-3D. Figure \ref{clusters} represents the faces of AFLW2000-3D in the pose space, where each face is colored according to the expert CNN with the most weight in the committee. Since a unique color is given for each expert CNN, we can observe the splitting performed by the gates on the dataset. Figure \ref{clusters} illustrates the repartition in head pose space from a Pose-Tree-MoE model trained on 300W-LP. We can then observe that on the first level of the tree, the red axis representing the yaw allows to separate the data associated with the two subtrees respectively. The same can be said for the second tree level, therefore the model learned to split the head pose space according to the yaw orientation primarily. This is consistent with the fact that the model was trained with 300W-LP, whose images essentially augmented with yaw.

\subsubsection{Regression layer}
\begin{figure}[!ht]
	\centering
	\includegraphics[width=\linewidth]{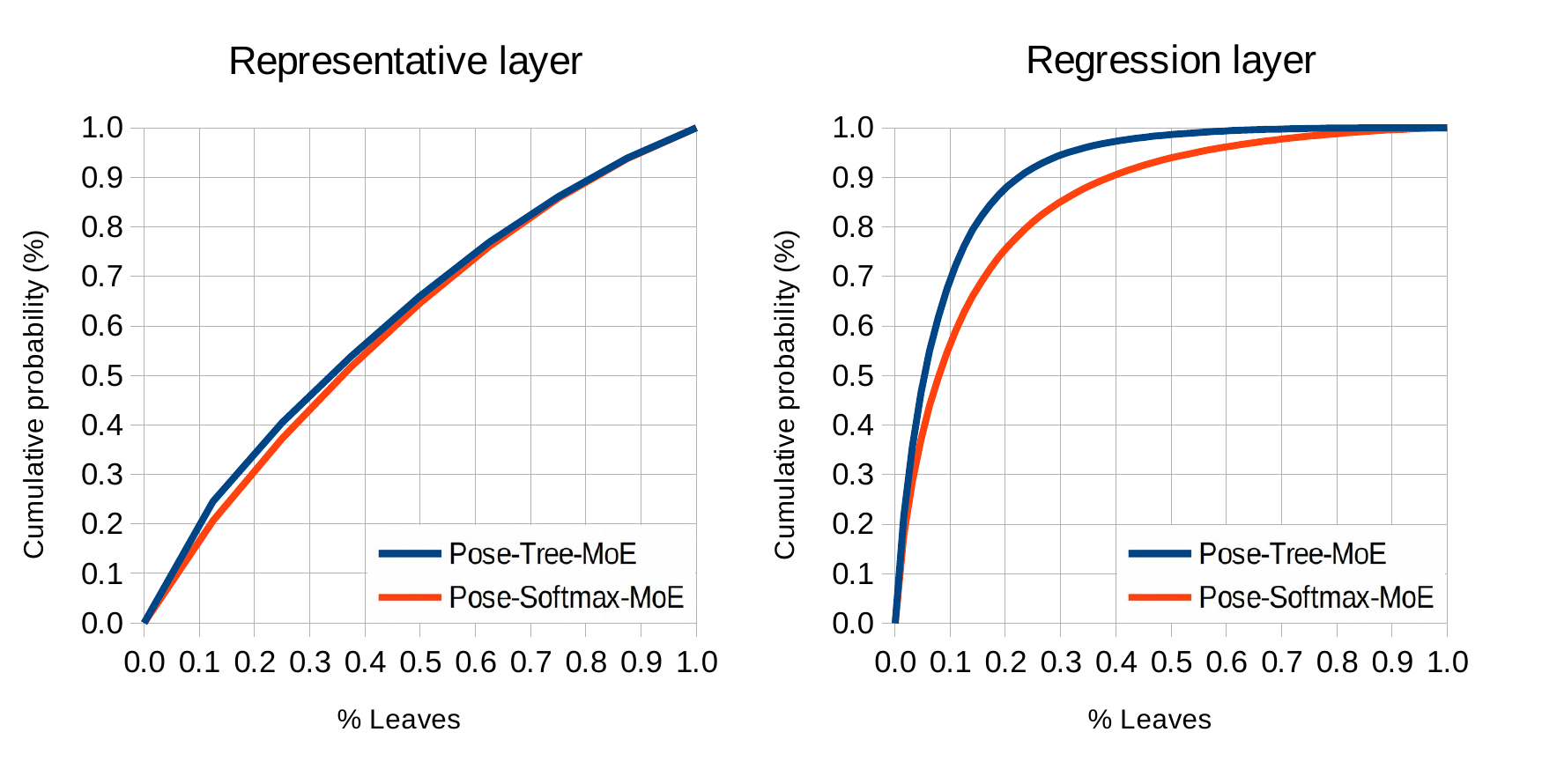}
	\caption{Cumulative top-scoring expert networks distribution for representation and regression layer at the first stage in the cascade for both Pose-Softmax-MoE and Pose-Tree-MoE.}
	\label{cumulated}
\end{figure}

\begin{figure*}[!ht]
	\centering
	\includegraphics[width=\linewidth]{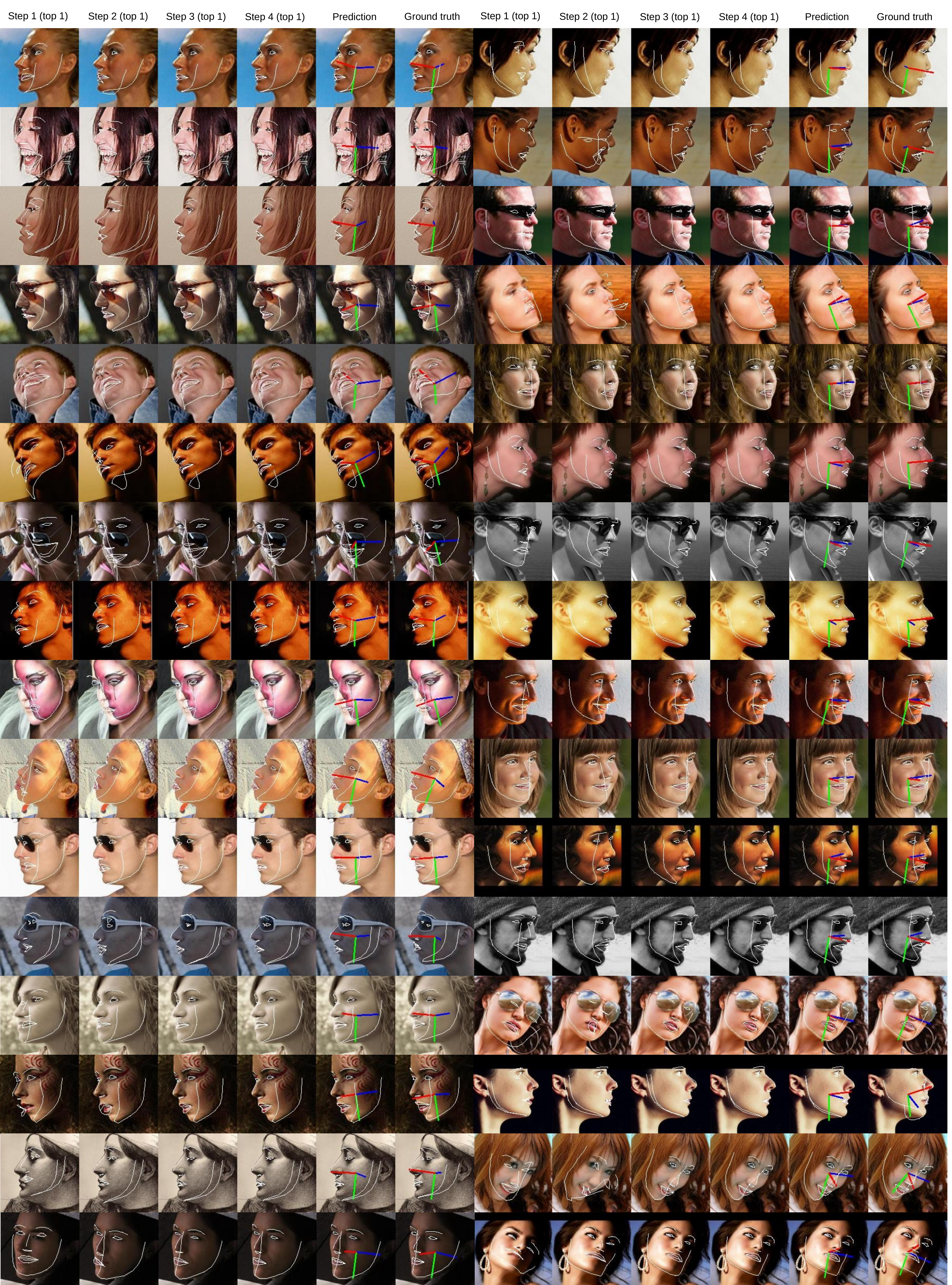}
	\caption{Visualisations of the predictions outputted for each cascade stage with only the top (maximum value of tree-gate) regressor. Head pose estimation is also displayed, as well as the ground truth. Images from AFLW2000-3D.}
	\label{displacements}
\end{figure*}

Figure \ref{cumulated} shows the average of the cumulative sum of the gates probabilities of expert CNNs, sorted in descending order on I-BUG. Notice on the right part that the tree-gated model allows $20\%$ of the regressors to explain more than $90\%$ of the final prediction, while the softmax-gated model needs about $40\%$ of the regressors to explain $90\%$ of the prediction. Thus, tree-gates allows to output a correct alignment using less expert regressors: the better repartition of expert regressors towards the specific alignment cases makes it possible to better specialize each expert regressor, and to use fewer of these to obtain a better representation. The results for the representation layer are less clear-cut, the Pose-Tree-MoE model lying marginally above the Pose-Softmax-MoE model. This is likely due to the lower number of experts (8 vs 64 for representation vs prediction), indicating that the difference between tree and softmax-gated MoE models becomes more conspicuous as the number of experts increases. All in all, tree gates promotes a more efficient repartition of the experts, and a better specialization thereof, which, in turns, leads to an overall higher accuracy and robustness.

\subsubsection{Visualizations}

Figure \ref{displacements} shows the predictions of our model ($5^{th}$ and $11^{th}$ columns) for large pose examples on AFLW2000-3D database, as compared with the ground truth markup ($6^{th}$ and $12^{th}$ columns). The prediction and ground truth landmark localizations are also plotted along with the estimated and ground truth head pose. For most examples, the head pose estimate is very close to its ground truth counterpart: this allows to select relevant expert CNNs in the representation layers, which give rise to high-quality landmark alignment even on examples exhibiting large pose variations.

Furthermore, rows 1 to 4 and 7 to 10  shows the displacements outputted by only one top-scoring regressor (as indicated by the associated tree-gate value), from the current shape at the corresponding cascade iteration. It should be noted that using a single regressor, our Pose-Tree-MoE model can achieve reasonable alignment accuracy, which will justify investigating the use of a restricted (top-k) numbers of experts in future work, e.g. using greedy evaluation as in \cite{Dapogny2018a}.

\subsubsection{Runtime evaluation}\label{runtime}
Last but not least, our method is very fast as it operates at $17.54$ ms per image on a NVIDIA GTX 1080 GPU, and thus can run at 57 fps. Furthermore, In \cite{shazeer2017outrageously}, MoE are used to reduce the computational load by keeping only a small number (top-k) of experts. With hierarchical gates, an interesting direction would be to evaluate the tree-gate in a greedy layerwise fashion as in \cite{Dapogny2018a}, and keep only the regressor corresponding to the maximum probability leaf to further reduce the computational cost.

\section{Conclusion}
In this paper, we have proposed to integrate ensemble methods within a deep architecture in order to increase the overall robustness of the model to large variations in the data. The use of a committee of experts neural networks instead of a single one allows an overall greater robustness. Furthermore, we showed that using a gate function to weight the responses of each expert network allows each of these networks to be more expert for a given context. In particular, the use of tree-gates makes it possible to jointly learn a committee of expert networks and a hierarchical clustering of the use of these experts. Additionally using neural trees to model the tree-gates allows to learn both the ensemble and associated gating network in an end-to-end manner.

As such, we showed that tree-gated MoE models can be used for modelling the regressors as well as the  feature representation layers, by using high-level semantic information such as head pose as a proxy variable. These tree-gates allows a more efficient clustering and specialization of the experts, leading to a higher performance. Furthermore, thorough experimental validation, we demonstrated that, when applied for face alignment in the frame of cascaded regression, the proposed approach yields high accuracies, most notably on challenging data in term of head pose and occlusion, while keeping a reasonable computational cost.

As a future work, we will investigate the use of a limited number (top-k) of experts for both the representation and regression layers, e.g. using greedy evaluation \cite{Dapogny2018a}, in order to further decrease the runtime. Furthermore, the tree-MoE architecture introduced in this paper is very generic and could be applied to a wide range of other computer vision problem, such as image classification, semantic segmentation, or object detection.

\section*{Acknowledgment}

This work has been supported by the French National Agency (ANR) in the frame of its Technological Research JCJC program (FacIL, project ANR-17-CE33-0002).

{
\bibliographystyle{ieee}
\bibliography{biblio}
}

\end{document}